\documentclass[letterpaper, 10 pt, conference]{IEEEtran}  % Comment this line out if you need a4paper

\IEEEoverridecommandlockouts                              % This command is only needed if 
\usepackage{graphics} % for pdf, bitmapped graphics files
\usepackage{subfigure}
\usepackage{newtxtext,newtxmath} % replacement for mathptx
\usepackage{times} % assumes new font selection scheme installed
\usepackage{balance}
\usepackage{amsmath} % assumes amsmath package installed
\usepackage{amssymb}  % assumes amsmath package installed
\usepackage{bm}
\usepackage{mathtools}
\usepackage{calc}
\usepackage{tensor}
\usepackage{etoolbox}
\undef{\t}
\undef{\v}
\usepackage{booktabs}
\usepackage{pifont}% http://ctan.org/pkg/pifont

\usepackage{todonotes}
\usepackage{soul}

\newcommand{\figref}[1]{Fig.~\ref{#1}}

\newcommand{\super}[1]{^{\scriptscriptstyle{\mathit{#1}\!}}}
\newcommand{\sub}[1]{_{\scriptscriptstyle{\!\mathit{#1}}}}
\newcommand{\minus}[1][]{\text{--}#1}

\newcommand{\MATRIX}[4]{%
	\ifstrempty{#2}
		{{ #1\super{#4}\sub{#3} }}
		{{ \super{#2\,}\bm{#1}\super{#4}\sub{#3} }}
}

\newcommand{\M}[3][] {\MATRIX{#2}{#1}{#3}{}}
\newcommand{\MT}[3][]{\MATRIX{#2}{#1}{#3}{\,\bf{T}\;}}
\newcommand{\MI}[3][]{\MATRIX{#2}{#1}{#3}{\,\bf{\minus1}\;}}

\newcommand{\A}[1][]{\M[\empty]{A}{#1}}
\newcommand{\AT}[1][]{\MI[\empty]{A}{#1}}
\newcommand{\B}[1][]{\M[\empty]{B}{#1}}
\newcommand{\BT}[1][]{\MI[\empty]{B}{#1}}

\newcommand{\E}[1][]{\M[\empty]{E}{#1}}

\newcommand{\X}{\M[\empty]{X}{\empty}}
\newcommand{\Y}{\M[\empty]{Y}{\empty}}

\newcommand{\T}[2][]{\M[#1]{T}{\!#2}}

\newcommand{\R}[2][]{\M[#1]{R}{#2}}

\newcommand{\V}[2][]{\M{\check{#2}}{#1}}

\newcommand{\v}[2][]{\M{\vec{#2}}{#1}}

\newcommand{\vT}[2][]{\MT{\vec{#2}}{#1}}

\newcommand{\t}[1]{\v[#1]{t}}

\newcommand{\tT}[1]{\vT[#1]{t}}

\newcommand{\q}[2][]{\M{\hat{#2}}{#1}}

\newcommand{\tightoverset}[2]{%
	\overset{\mathrel{\raisebox{-0.5ex}{$#1$}}}{#2}
}

\newcommand{\mat}[2]{\M{\tightoverset{\scriptstyle{\bm{#2}}}{[{#1}]}}{}}
\newcommand{\matT}[2]{\M{\tightoverset{\scriptstyle{\!\!\!\bm{#2}}}{[{#1}]\super{\bf{T}}}}{}}

\newcommand{\MP}[1]{\mat{#1}{+}}

\newcommand{\MM}[1]{\mat{#1}{-}}
\newcommand{\MMT}[1]{\matT{#1}{-}}

\newcommand{\qM}[3][]{\mat{{\q[#1]{#2}}}{#3}}
\newcommand{\qMT}[3][]{\matT{{\q[#1]{#2}}}{#3}}

\newcommand{\qMP}[2][]{\qM[#1]{#2}{+}}
\newcommand{\qMPT}[2][]{\qMT[#1]{#2}{+}}

\newcommand{\qMM}[2][]{\qM[#1]{#2}{-}}
\newcommand{\qMMT}[2][]{\qMT[#1]{#2}{-}}

\newcommand{\vM}[3][]{\mat{{\v[#1]{#2}}}{#3}}
\newcommand{\vMT}[3][]{\matT{{\v[#1]{#2}}}{#3}}

\newcommand{\vMP}[2][]{\vM[#1]{#2}{+}}
\newcommand{\vMPT}[2][]{\vMT[#1]{#2}{+}}

\newcommand{\vMM}[2][]{\vM[#1]{#2}{-}}
\newcommand{\vMMT}[2][]{\vMT[#1]{#2}{-}}

\newcommand{\kron}[2]{#1\otimes#2}
\newcommand{\kronVec}[1]{\text{vec}(#1)}

\newcommand{\thead}[1]{\bfseries{#1}}

\title{\LARGE Continuous hand-eye calibration using 3D points}

\author{%
\IEEEauthorblockN{Bjarne Grossmann}
\IEEEauthorblockA{Robotics, Vision, and Machine Intelligence\\
Aalborg University Copenhagen, Denmark\\
bjarne@m-tech.aau.dk}
\and
\IEEEauthorblockN{Volker Kr\"uger}
\IEEEauthorblockA{Robotics, Vision, and Machine Intelligence\\
Aalborg University Copenhagen, Denmark\\
vok@m-tech.aau.dk}
}

\begin{document}

\maketitle
\thispagestyle{empty}
\pagestyle{empty}

\begin{abstract}

The recent development of calibration algorithms has been driven into two major directions:
(1) an increasing accuracy of mathematical approaches and (2) an increasing flexibility in usage by reducing the dependency on calibration objects.
These two trends, however, seem to be contradictory since the overall accuracy is directly related to the accuracy of the pose estimation of the calibration object and therefore demanding large objects, while an increased flexibility leads to smaller objects or noisier estimation methods.

The method presented in this paper aims to resolves this problem in two steps:
First, we derive a simple closed-form solution with a shifted focus towards the equation of translation that only solves for the necessary hand-eye transformation.
We show that it is superior in accuracy and robustness compared to traditional approaches.
Second, we decrease the dependency on the calibration object to a single 3D-point by using a similar formulation based on the equation of translation which is much less affected by the estimation error of the calibration object's orientation.
Moreover, it makes the estimation of the orientation obsolete while taking advantage of the higher accuracy and robustness from the first solution, resulting in a versatile method for continuous hand-eye calibration.

\end{abstract}

\section{INTRODUCTION}\label{sec:introduction}

The coordination between robot eyes (visual sensor) and a hand (endeffector) is crucial for executing vision-based tasks.
This coordination can be expressed as a spatial relationship between the sensor and the endeffector.
Determining this rigid transformation is referred to as robot-sensor or hand-eye calibration and can be computed by moving the robot while observing the motion with the sensor.
%This usually involves a pose estimation of a large calibration object which makes continuous calibration unfeasible.
\begin{figure}[!ht]
	\centering
	\subfigure[][Eye-on-hand]{
		\includegraphics[width=0.45\columnwidth]{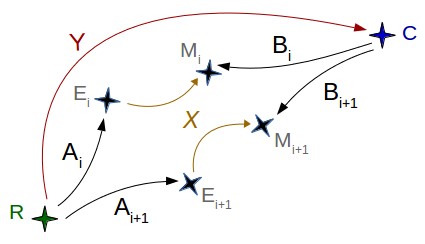}
		\label{fig:EyeOnHand}
	}
	\subfigure[][Eye-in-hand]{
		\includegraphics[width=0.45\columnwidth]{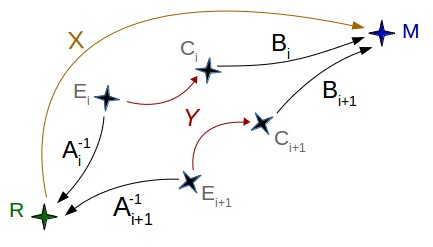}
		\label{fig:EyeInHand}
	}
	\caption{shows the two different setups defined by the calibration problem. Depicted are two measurements of the \textbf{R}obot - \textbf{E}ndeffector transformation \textbf{A} and the \textbf{C}amera - \textbf{M}arker transformation \textbf{B} respectively. \textbf{X} and \textbf{Y} are unknown and have to be estimated.}
	\label{fig:setup}	
\end{figure}

Mathematically, the relationship can in general be written as an equation of homogeneous transformation matrices of the form $\bm{AX} = \bm{YB}$ (cf. \figref{fig:setup}) which can be solved for $\bm{X}$ and $\bm{Y}$ simultaneously.
However, $\bm{X}$ and $\bm{Y}$ can also be computed sequentially using two measurements and rearranging it to an equation of form $\bm{A^{\!\scriptscriptstyle\prime}X} = \bm{XB^{\scriptscriptstyle\prime}}$ (cf. sec.~\ref{sec:problem}).

Several closed-form / linear least-square solutions have been derived in the past decades to solve these equations.
They have two characteristics in common:
First, they solve the whole equation system including the transformation of the calibration object with respect to the robot.
Second, to solve the equations, they require an accurate pose estimation of the calibration object from the sensor data.

For those reasons, traditional solutions rely heavily on large calibration objects like chessboards or similar whose pose can be easily estimated, here, the accuracy increases with the size of the object.
This yields an inflexible calibration procedure that has to be run during a separate setup phase of the robot making it a time-consuming and fiddly task.
In real-world applications, particularly in industrial scenarios, this calibration process is often not feasible as it interrupts the actual production workflow.
Recalibration of the sensor is especially necessary on mobile robot platforms that tend to lose calibration due to vibrations while moving.

Methods to reduce the use of calibration objects and using smaller markers, planes or even only single points have been suggested \cite{1992-Wang, 2002-Zhang, 2014-Hu, 2014-Lundberg}, but require new motion constraints on the robot which makes these methods unfeasible for continuous calibration.
In the recent years, structure-from-motion approaches \cite{2001-Andreff, 2011-Heller} have been developed making a calibration object obsolete, but they rely heavily on 2D image feature detectors and correspondence matching or tracking to find the homography between two point sets (basically using the world as a marker).
Even though the results are promising for eye-in-hand calibration, the opposite case, using a static camera is more difficult to solve, as multiple points belonging to the endeffector have to be detected and matched.

This paper presents a novel view on the calibration problem based on available closed-form solutions to compute the hand-eye transformation based on the assumption that the calibration object is solely a means to an end.
The paradigm behind our approach is to not solve the whole equation system and consider the solution of the orientation related to the calibration object as unnecessary.
We first formulate a quaternion-based approach directly derived from existing closed-form solutions, but solving for less unknowns and show that the system is still solvable for the eye-robot transformation resulting in solutions of higher accuracy and robustness.
Afterwards, we derive an equivalent formulation using the Kronecker product which allows us to reduce the calibration object to a single point without constraining the robot movement and therefore making the pose estimation of the calibration object obsolete.
Utilizing this formulation enables us to define a new calibration procedure with (1) the high accuracy of the reformulated closed-form solution, (2) no need for a calibration object except for a single point, making it a highly versatile method and thereby (3) allowing a continuous calibration during the robot's actual tasks.

In sec.~\ref{sec:background}, the calibration problem is defined followed by the classical solution process using the quaternion representation or the Kronecker product.
Here, we also define the basic notation which is used throughout the paper.
Section \ref{sec:theory} picks up on the theory and formulates the problem in a new, more general way which allows us to structure the possible solution space for the calibration problem.
As a result, we are able to find "missing", or unknown solutions where especially two of them lead to our proposed solution in sec.~\ref{sec:proposed}.
We describe the derivation of our approach here and finally prove its validity by conducting comparative experiments in sec.~\ref{sec:experiments} followed by a conclusion in sec.~\ref{sec:conclusion}.

\section{BACKGROUND \& THEORY}\label{sec:background}

\subsection{Problem statement}\label{sec:problem}
The spatial relationship between robot $\bm{R}$, endeffector $\bm{E}$, camera sensor $\bm{C}$ and marker (calibration object) $\bm{M}$ in a "eye-on-hand" calibration scenario (static camera with marker on the endeffector) can be seen in \figref{fig:EyeOnHand}.
Note that by exchanging the sensor and the marker, the same formulation can be used for the "eye-in-hand" calibration (endeffector-mounted camera and static marker as world frame) as depicted in \figref{fig:EyeInHand}.
The calibration problem can be derived and written as
\begin{align}
	\A[i] \X &= \Y \B[i] \label{eq:tf:XY} \\
	\left( \AT[i+1] \A[i] \right) \X &= \X \left( \BT[i+1] \B[i] \right) \label{eq:tf:X} \\
	\left( \A[i] \AT[i+1] \right) \Y &= \Y \left( \B[i] \BT[i+1] \right) \label{eq:tf:Y}
\end{align}
where $\A[i] = \T[R]{E_i}$ denotes the transformation matrix between the robot base and an arbitrary pose $\E[i]$ of the endeffector and $\B[i] = \T[C]{M_i}$ the transformation between the camera and the observed marker for the $i_{\text{th}}$ measurement.
Obviously, \eqref{eq:tf:X} and \eqref{eq:tf:Y} are instances of the general problem formulation $\bm{AX} = \bm{XB}$ which in turn is a particular instance of $\bm{AX} = \bm{YB}$ for $\bm{Y} = \bm{X}$.
Therefore, a solution to $\bm{AX} = \bm{XB}$ is just a simpler version of the more general problem \eqref{eq:tf:XY}.
Since in that case $\bm{A}$ and $\bm{B}$ are similar, it follows that the matrices have the same eigenvalues and the same angle of rotation with respect to their rotation axes, which helps to simplify the solutions even more.
In this paper, we focus on the general case $\bm{AX} = \bm{YB}$.

\subsection{General Solution}\label{sec:solutions}
The general approach to solve the calibration equation is to reformulate it as a (homogeneous) linear equation system.
Since in most applications \eqref{eq:tf:XY} is underdetermined and subject to noise, multiple measurements, at least three (cf. \cite{1994-Zhuang, 2013-Shah}), of $A$ and $B$ are necessary.

The classical methods split \eqref{eq:tf:XY} into two parts and solve them sequentially by first computing the rotation and then using it to find the according translation.
This can be achieved by writing the transformation matrix $T$ as a composition of its rotational part $\R{T}$ and translational part $\t{T}$ and thus express \eqref{eq:tf:XY} as
\begin{align*}
	\begin{bsmallmatrix} \R{A} & \t{A} \\ \vT{0} & 1 \end{bsmallmatrix}
	\begin{bsmallmatrix} \R{X} & \t{X} \\ \vT{0} & 1 \end{bsmallmatrix}
	=
	\begin{bsmallmatrix} \R{Y} & \t{Y} \\ \vT{0} & 1 \end{bsmallmatrix}
	\begin{bsmallmatrix} \R{B} & \t{B} \\ \vT{0} & 1 \end{bsmallmatrix}
\end{align*}
with its decomposition into two separate equations
\begin{align}
	\R{A}\R{X} &= \R{Y}\R{B} \label{eq:R} \\
	\R{A}\t{X} + \t{A} &= \R{Y}\t{B} + \t{Y} \label{eq:t}
\end{align}
and where \eqref{eq:R} is known as the equation of rotation and \eqref{eq:t} as the equation of translation.
Since once the equation of rotation is solved and inserted into the translational part, it can be easily reduced to a linear least-square problem:
\begin{align}
	\begin{bsmallmatrix} \R{A} & \minus{I} \end{bsmallmatrix}
	\begin{bsmallmatrix} \t{X} \\ \t{Y} \end{bsmallmatrix}
	= \R{Y}\t{B} - \t{A}
\end{align}
However, the rotational part \eqref{eq:R} is not directly solvable with respect to $\bm{X}$ and/or $\bm{Y}$ due to the the noncommutativity of the matrix multiplication.

In the last decades, the solution to this problem has been extensively researched and several solutions have been proposed to solve the equation of rotation \eqref{eq:R}.
Most approaches are based on rephrasing the problem using a different representation for the rotation matrices including quaternions or the Kronecker product as the most prominent ones.

However, Chen \cite{1991-Chen} argues that the sequential solution of the robot-sensor calibration is not valid, since it implies that the rotational and translational part are independent from each other.
Therefore, simultaneous solutions have been developed which combine both parts into one equation system by rearranging the equation of translation \eqref{eq:t}:
\begin{align}
	\R{Y}\t{B} + \t{Y} - \R{A}\t{X} = \t{A} \label{eq:solve:t}
\end{align}

In the following, we give an overview of sequential and simultaneous solutions in the notation of quaternions and the Kronecker product.
The notation is then used to generalize the representation-specific solutions which allows us to define the solution space more systematically and thereby find unknown methods.

\subsection{Quaternions}\label{sec:quaternion}

A unit quaternion is denoted as
\begin{align*}
	\q{q} = \begin{bsmallmatrix} \M{q}{w} \\ \v[v]{q} \end{bsmallmatrix}
	      = \begin{bsmallmatrix} \cos{\frac{\alpha}{2}} \\ \sin{\frac{\alpha}{2}}\v[e]{q} \end{bsmallmatrix}
	      = \M{q}{w} \begin{bsmallmatrix} 1 \\ \tan{\frac{\alpha}{2}}\v[e]{q} \end{bsmallmatrix} 
%	      = \M{q}{w} \left(\begin{bsmallmatrix} 1 \\ \v{0} \end{bsmallmatrix} + \q[0]{q}\right)
\end{align*}
where $\v[e]{q}$ denotes the normalized Euler vector or rotation axis and $\alpha$ the rotation angle.
The quaternion multiplication is defined as
\begin{align*}
	\q{p}\q{q} = \begin{bsmallmatrix} \M{p}{w}\M{q}{w} - \vT[v]{p}\v[v]{q}\\
				 \v[v]{p}\M{q}{w} + (\M{p}{w}I + \vMP{p})\v[v]{q} \end{bsmallmatrix}
			   = \begin{bsmallmatrix} \M{q}{w}\M{p}{w} - \vT[v]{q}\v[v]{p}\\
			   	 \v[v]{q}\M{p}{w} + (\M{q}{w}I + \vMM{q})\v[v]{p} \end{bsmallmatrix}
\end{align*}
with $\vMP[v]{q}$ and $\vMM[v]{q}$ denoting the cross-product matrix
\begin{align*}
	\vMP[v]{q} = \minus{}\vMM[v]{q} = \minus{}\vMPT[v]{q} = \vMMT[v]{q}
	= \begin{bsmallmatrix} 0 & \minus{}\M{q}{v3} & \M{q}{v2}\\
						   \M{q}{v3} & 0 & \minus{}\M{q}{v1}\\
						   \minus{}\M{q}{v2} & \minus{}\M{q}{v1} & 0
	\end{bsmallmatrix} \enskip.
\end{align*}
Factoring out $\q{p}$ or $\q{q}$, we get
\begin{align*}
	\q{p}\q{q} &= \begin{bsmallmatrix} \M{p}{w} & \minus{}\vT[v]{p}\\
				                      \v[v]{p} & \ \M{p}{w}I + \vMP{p} \end{bsmallmatrix}
				 \begin{bsmallmatrix} \M{q}{w} \\ \\ \v[v]{q} \end{bsmallmatrix}
			   = \begin{bsmallmatrix} \M{q}{w} & \minus{}\vT[v]{q}\\
			   	 \v[v]{q} & \ \M{q}{w}I + \vMM{q} \end{bsmallmatrix}
				 \begin{bsmallmatrix} \M{p}{w} \\ \\ \v[v]{p} \end{bsmallmatrix}
\end{align*}
which shows that the quaternion multiplication is commutative up to the sign of the cross-product matrix.
A quaternion multiplication and therefore a concatenated rotation can now be expressed in terms of the matrix multiplication
\begin{align}
	\q{p}\q{q} = \qMP{p}\q{q} = \q{q}\qMPT{p} = \q{p}\qMMT{q} = \qMM{q}\q{p} \label{eq:Q:mapping} \enspace.
\end{align}
Furthermore, Horn \cite{1987-Horn} showed that \eqref{eq:R} can be expressed in terms of quaternions and is equivalent to solving $\q{a}\q{x} = \q{y}\q{b}$ which therefore becomes
\begin{align}
	\qMP{a}\q{x} = \qMM{b}\q{y} \enspace. \label{eq:QR}
\end{align}
This formula has been adapted and led to solutions derived and implemented by \cite{1991-Chou} for the $AX=XB$ case and \cite{1998-Dornaika} for the $AX=YB$.

Since the axis-angle representation is closely related to the quaternions, as seen in the quaternion definition, classical approaches for solving the $AX=XB$ equation like \cite{1989-Tsai, 1992-Wang, 1994-Park} can be easily derived which otherwise have to be proven in a much more complicated manner, e.g. through geometry, group or screw theory.
Zhuang \cite{1994-Zhuang} showed that the axis-angle approach can be extended for solving also the $AX=YB$ equation by inserting the scalar part of a quaternion into the imaginary part and thereby reducing the dimensionality to 3D. 

Furthermore, the quaternion representation can also be applied to the equation of translation \eqref{eq:solve:t} as shown by \cite{1995-Lu, 1999-Daniilidis} using different derivations like 8-space quaternions or dual quaternions for solving the calibration problem for rotation and translation simultaneously.
The equation can be written as
\begin{align}
	\begin{bsmallmatrix} \qMP[A]{t}\qMP{a} \:&\: \qMP{a} \:&\: \minus{}\qMM{b}\qMM[B]{t} \:&\: \minus{}\qMMT{b} \end{bsmallmatrix}
	\begin{bsmallmatrix} \q{x} \\ \qMM{x}\q[X]{t} \\ \\ \q{y} \\ \qMM{y}\q[Y]{t} \end{bsmallmatrix}
	= 0 \label{eq:solve:tQ:1}
\end{align}
where $\q[Q]{t}$ represents the pure imaginary quaternion of $\t{Q}$.
This equation is sufficient to solve for $\X$ and $\Y$, is however usually combined with the equation of rotation to inject an additional constraint by simply stacking the equations.
The problem about the solution space of \eqref{eq:solve:tQ:1} are the pure imaginary quaternions which scalar part cannot be constrained to be $0$, resulting in an underdetemined equation system with a nullity of two.
The authors in \cite{1999-Daniilidis, 1995-Lu} showed however that the equation is still solvable, since the solution is a linear combination of the two singular vectors (corresponding to the two vanishing singular values) spanning the null space.
$\q[X]{t}$ and $\q[Y]{t}$ can then be retrieved by postmultiplying with $\scriptstyle\qMM{x}$ and $\scriptstyle\qMM{y}$ respectively, since $\q[Q]{t} = {\scriptstyle\qMM{q}}\q[Q]{t}{\scriptstyle\qMM{q}} = {\scriptstyle\qMM{q}\qMMT{q}}\:\q[Q]{t} = \q[Q]{t}$.
\newline

\subsection{Kronecker product}\label{sec:kronecker}
The Kronecker product $\otimes$ of two matrices $A$ and $B$ is defined as
\begin{align*}
	\kron{A}{B} = \begin{bsmallmatrix}
		\M{a}{11}B & \cdots & \M{a}{1n}B \\
		\vdots     & \ddots & \vdots \\ \\
		\M{a}{m1}B & \cdots & \M{a}{mn}B
	\end{bsmallmatrix} \enskip.
\end{align*}
An interesting property of the Kronecker product which allows us to write a matrix equation $\M{Z}{} = \M{A}{}\M{X}{}\M{B}{}$ in a more convenient form, is given by
\begin{align}
	\kronVec{\M{Z}{}} = \kronVec{\M{A}{}\M{X}{}\M{B}{}} = (\kron{\MT{B}{}}{\M{A}{}})\:\kronVec{\M{X}{}} \nonumber
\end{align}
where $\kronVec{}$ denotes the vectorization of a matrix into a single vector.
More generally, we can state that a matrix multiplication $AB$ can be expressed as
\begin{align}
	\begin{split}
		\kronVec{\M{A}{}\M{B}{}} &= (\kron{\M{I}{}}{\M{A}{}})\:\kronVec{\M{B}{}}  = \kronVec{\M{B}{}}(\kron{\M{I}{}}{\MT{A}{}}) \\
		   &= \kronVec{\M{A}{}}(\kron{\M{B}{}}{\M{I}{}}) = (\kron{\MT{B}{}}{\M{I}{}})\:\kronVec{\M{A}{}} \enskip. 
	\end{split} \label{eq:K:mapping}
\end{align}

This property leads straight forward to the rephrased formulation of \eqref{eq:R} as
\begin{align}
	(\kron{\M{I}{}}{\M{A}{}})\:\kronVec{\M{X}{}} &= (\kron{\MT{B}{}}{\M{I}{}})\:\kronVec{\M{Y}{}}
\end{align}
which has been first applied to the calibration problem by \cite{2001-Andreff} for the $AX=XB$ case.
The Kronecker product has been a base for further implementations, extensions and simplifications in \cite{2008-Liang, 2010-Li, 2013-Shah}.

The straight-forward application of the Kronecker product makes it also quite easy to rephrase the equation of translation \eqref{eq:t} making it possible to solve the calibration problem simultaneously.
The equation becomes
\begin{align}
	[\kron{\tT{B}}{I}]vec(\R{Y}) + \t{Y} - \R{A}\t{X} = \t{A} \label{eq:solve:tK:1} \enskip.
\end{align}
Stacking this formulation with the equation of rotation, we can solve for the whole system in one step as shown by \cite{2001-Andreff, 2010-Li}.
Note that the solution does not obey the orthonormality constraint of rotation matrices and hence has to be reorthonormalized.
However, the corrected error is not reflected in the translational part which has to be recomputed in a second step by solving the equation of translation with the now known rotation again.

\section{PROPOSED GENERALIZATION OF THE THEORY}\label{sec:theory}

In the following section, we analyze existing approaches for both representations and find an abstract solution formulation for the calibration problem.
It allows us to give a more structured definition of the solution space which we utilize to represent existing and derive new methods.

Looking at the way how the equation of rotation \eqref{eq:R} is reformulated in quaternion representation or using the Kronecker product, we see that the general solution to the problem is to rephrase a plain matrix multiplication into a commutative matrix-vector multiplication.
The key properties to achieve this are given in \eqref{eq:Q:mapping} and \eqref{eq:K:mapping}.

We suggest that both properties are represented by a general linear mapping $\Omega$ such that
\begin{align}
	\Omega(AB) = \MP{A}\V{b} = \MM{B}\V{a} \label{eq:mapping}
\end{align}
with $\MP{\,}$ and $\MM{\,}$ denoting a matrix representation for left and right multiplication and $\V{\,}$ a vectorized representation of the respective input.

Therefore, the commutation rules of the different representation can be summarized as
\begin{align}
	\begin{aligned}
		\Omega(AB)
		&= \MP{A}\V{b}  &\enskip\Leftrightarrow\enskip& \qMP{a}\q{b}  &\enskip\Leftrightarrow\enskip& (\kron{\M{I}{}}{\M{A}{}})\:\kronVec{\M{B}{}} \\
		&= \MM{B}\V{a}  &\enskip\Leftrightarrow\enskip& \qMM{a}\q{b}  &\enskip\Leftrightarrow\enskip& (\kron{\MT{B}{}}{\M{I}{}})\:\kronVec{\M{A}{}} \enskip.
	\end{aligned}
\end{align}
which obviously holds for both properties.

Using this mapping, \eqref{eq:R} can be rewritten in two different solvable equations.
The first approach is to directly solve
\begin{align}
	\left[ \MP{\R{A}} \enskip \minus{}\MM{\R{B}} \right] \begin{bsmallmatrix} \V[X]{r} \\ \V[Y]{r} \end{bsmallmatrix}
	=
	\left[ \MMT{\R{B}}\MP{\R{A}} \enskip \minus{}I \right] \begin{bsmallmatrix} \V[X]{r} \\ \V[Y]{r} \end{bsmallmatrix}
	= 0 \label{eq:solve:R:1}
\end{align}
which can further simplified (cf.~\cite{1998-Dornaika, 2013-Shah}) and is equivalent to decomposing
\begin{align}
	K = \sum\limits_{i=1}^n{\left(\MMT{\R{B_i}}\MP{\R{A_i}}\right)} \label{eq:solve:R:3}
\end{align}
using the SVD with $\V[X]{r}$ and $\V[Y]{r}$ being proportional to the right and left singular vector corresponding to the highest eigenvalue of $K$, respectively.

The second approach formulates \eqref{eq:R} as a simple orthogonal Procrustes problem (cf.~\cite{1994-Park}) given as
\begin{align}
	 \left( \MMT{\R{X}} \MP{\R{Y}}\right) \V[B]{r} = \V[A]{r} \label{eq:solve:R:2}
\end{align}
and can also be solved by the SVD where the multiplication of the left and right singular matrix results in the unknown matrix $(\MMT{\R{X}} \MP{\R{Y}})$.
\newline

These general formulations let us describe the solution space for the calibration problem more systematically by either using the quaternion representation or the Kronecker product in \eqref{eq:solve:R:1}, \eqref{eq:solve:R:2} or \eqref{eq:solve:R:3}.
Additionally, \eqref{eq:solve:R:1} can also be combined with \eqref{eq:solve:tQ:1} or \eqref{eq:solve:tK:1} depending on the representation resulting in simultaneous solutions.
An excerpt of feasible solutions derived this way leading to well known and new ones is shown in Tbl.~\ref{tbl:implementations}.
We furthermore include solutions based on the axis-angle representations since they can be easily derived from the quaternion formulation.

As we can see, solutions solely based on the equation of translation have, to our best knowledge, not been developed in the past, they usually are part of the simultaneous solutions though.
On the one hand, this might be due to the fact that \eqref{eq:solve:tK:1} does only allow to estimate one rotation, and on the other hand, that they tend to be less stable when used stand-alone as formulated in \eqref{eq:solve:tQ:1}.

\begin{table}[tb]
	\vspace{7pt}
	\centering
	\resizebox{\columnwidth}{!}{
	\small
	\begin{tabular}{lllll}
	 	\toprule
	 	\thead{TYPE} & \thead{REPR.} & \thead{EQ.} & \thead{AUTHOR} & \thead{NAME} \\
	 	\midrule
	 	AX=XB & AxisAngle & R  (\ref{eq:solve:R:1}) & Tsai~\cite{1989-Tsai}         & -- \\
	 	AX=XB & AxisAngle & R  (\ref{eq:solve:R:2}) & Park~\cite{1994-Park}         & XAxisRX \\
	 	AX=XB & Quaternion& R  (\ref{eq:solve:R:1}) & Chou~\cite{1991-Chou}         & -- \\
	 	AX=XB & Quaternion& R  (\ref{eq:solve:R:2}) & \textit{\textbf{this}}                 & -- \\
	 	AX=XB & Quaternion& R  (\ref{eq:solve:R:3}) & cf.~Dornaika~\cite{1998-Dornaika} & XQuatR$\ast$ \\
	 	AX=XB & Kronecker & R  (\ref{eq:solve:R:1}) & Liang~\cite{2008-Liang}       & -- \\
	 	AX=XB & Kronecker & R  (\ref{eq:solve:R:3}) & cf. Shah~\cite{2013-Shah}     & XKronR$\ast$ \\
	 	AX=XB & Kronecker & R  (\ref{eq:solve:R:2}) & \textit{\textbf{this}}                 & -- \\
	 	AX=XB & Quaternion& RT (\ref{eq:solve:R:1},\ref{eq:solve:tQ:1}) & Daniilidis~\cite{1999-Daniilidis} & -- \\
	 	AX=XB & Quaternion& RT (\ref{eq:solve:R:1},\ref{eq:solve:tQ:1})$^{\!\ast}$ & Lu~\cite{1995-Lu}                 & -- \\
	 	AX=XB & Kronecker & RT (\ref{eq:solve:R:1},\ref{eq:solve:tK:1}) & Andreff~\cite{2001-Andreff}       & XKronRT \\
	 	AX=XB & Quaternion& T  (\ref{eq:solve:tQ:1}) & \textit{\textbf{this}} & -- \\
	 	AX=XB & Quaternion& T  (\ref{eq:solve:tQ:1})$^{\!\ast}$ & \textit{\textbf{this}} & -- \\
	 	AX=XB & Kronecker & T  (\ref{eq:solve:tK:1}) & \textit{\textbf{this}} & XKronT \\
	 	\midrule
	 	AX=YB & AxisAngle & R  (\ref{eq:solve:R:1}) & Zhuang~\cite{1994-Zhuang}     & YAxisR \\
	 	AX=YB & Quaternion& R  (\ref{eq:solve:R:1}) & cf. Chou~\cite{1991-Chou}     & -- \\
	 	AX=YB & Quaternion& R  (\ref{eq:solve:R:3}) & Dornaika~\cite{1998-Dornaika} & YQuatR$\ast$ \\
	 	AX=YB & Kronecker & R  (\ref{eq:solve:R:1}) & cf. Liang~\cite{2008-Liang}   & -- \\
	 	AX=YB & Kronecker & R  (\ref{eq:solve:R:3}) & Shah~\cite{2013-Shah}         & -- \\
	 	AX=YB & Quaternion& RT (\ref{eq:solve:R:1},\ref{eq:solve:tQ:1}) & cf. Daniilidis~\cite{1999-Daniilidis} & -- \\
	 	AX=YB & Kronecker & RT (\ref{eq:solve:R:1},\ref{eq:solve:tK:1}) & Li~\cite{2010-Li} & YKronRT \\
	 	AX=YB & Quaternion& T  (\ref{eq:solve:tQ:1}) & \textit{\textbf{this}} & -- \\
	 	AX=YB & Quaternion& T  (\ref{eq:solve:this:Q}) & \textit{\textbf{this - proposed}} & YQuatT' \\
	 	AX=YB & Kronecker & T  (\ref{eq:solve:this:K}) & \textit{\textbf{this - proposed}} & YKronT' \\
		\bottomrule
	\end{tabular}
	}
	\caption{Implemented algorithms based on the general solution. Methods with names are used in the evaluation.
	\hspace{\columnwidth}$^*$\tiny{The formulation of the equation is slightly different.}}
	\label{tbl:implementations}
	\vspace{-20pt}
\end{table}

As an additional result of the analysis of the solution space, we implemented all solvers mentioned in Tbl.~\ref{tbl:implementations} in a easy-to-use python toolbox including a chessboard and an Aruco [http://sourceforge.net/projects/aruco] marker detector and an interface to ROS (Robot Operating System) [http://www.ros.org] which are available online [http://github.com/Bjarne-AAU/HandEyeCalibration].

\section{PROPOSED SOLUTIONS}\label{sec:proposed}

The analysis of the general solution in the previous section showed that the available solutions are grounded on the use of the equation of rotation alone or in combination with the equation of translation, however, no solutions based on the latter only have been investigated.
As a result, we use this gap as a foundation for the development of our proposed solutions and show in the following that the equation of translation alone is sufficient to solve the hand-eye calibration and in fact, results in an approach which is superior in convergence and accuracy compared to other methods.
The key aspect here is to shift from solving the whole equation system for all unknown parameters $\X$, $\Y$ to a reduced one that only estimates the necessary hand-eye transformation $\Y$.
Finally, we show that one particular solution enables us to not only accurately solve the calibration problem, but also does not rely on the estimated orientation of the calibration object, resulting in a simple, yet flexible method which can be used for continuous hand-eye calibration.

Starting from \eqref{eq:solve:tQ:1}, we apply the definition \eqref{eq:QR} and use the property \eqref{eq:Q:mapping} such that
\begin{align*}
 	&\qMM{b}\qMM[B]{t}\q{y} - \qMP[A]{t}\qMP{a}\q{x} + \qMM{b}\qMM{y}\q[Y]{t} - \qMP{a}\qMM{x}\q[X]{t} = 0 \\
 	\Rightarrow\enskip &\qMM{b}\qMM[B]{t}\q{y} - \qMP[A]{t}\qMP{y}\q{b} + \qMM{b}\qMM{y}\q[Y]{t} - \qMP{a}\qMM{x}\q[X]{t} = 0 \\
\end{align*}
\begin{align*}
 	\Rightarrow\enskip &\qMM{b}\qMM[B]{t}\q{y} - \qMM{b}\qMP[A]{t}\q{y} + \qMM{b}\qMM{y}\q[Y]{t} - \qMP{a}\qMM{x}\q[X]{t} = 0 \\
 	\Rightarrow\enskip &\left(\qMM[B]{t} - \qMP[A]{t}\right)\q{y} + \qMM{y}\q[Y]{t} - \qMMT{b}\qMP{a}\qMM{x}\q[X]{t} = 0 \\
\end{align*}
or in matrix form
\begin{align}
	\begin{bmatrix} \qMM[B]{t} - \qMP[A]{t} & I & \minus{}\qMMT{b}\qMP{a} \end{bmatrix}
	\begin{bmatrix} \q{y} \\ \qMM{y}\q[Y]{t} \\ \qMM{x}\q[X]{t} \end{bmatrix}
	= 0 \label{eq:solve:this:Q}
\end{align}
such that we do not explicitly solve for $\q{x}$ anymore and its matrix counterpart $\mat{\q{x}}{}$ can be ignored since we do not want to extract $\t{X}$ anyway.
Since this reformulation uses the same input as the original formulation but solves for less parameters, the least-square solution is therefore more accurate as shown in sec.~\ref{sec:experiments} and can be easily solved using the SVD.
Note that, as mentioned in section~\ref{sec:quaternion}, the equation is underdetermined such that $\q{y}$ and $\q[Y]{t}$ have to be computed as a linear combination of the two base vectors spanning the null space \cite{1999-Daniilidis}.

Even though we can achieve a more accurate estimation of the hand-eye transformation, the quaternion formulation is still dependent on the marker orientation $\q{b}$ and the error of its estimation.

However, taking \eqref{eq:solve:t} and representing it with the Kronecker product as in \eqref{eq:solve:tK:1}, we are able to tackle this problem which is given by
\begin{align}
	\begin{bmatrix} (\kron{\tT{B}}{I}) & I & \minus{}\R{A} \end{bmatrix}
	\begin{bmatrix} vec(\R{Y}) \\ \t{Y} \\ \t{X} \end{bmatrix}
	= \t{A} \enskip. \label{eq:solve:this:K}
\end{align}
In contrast to the quaternion approach, \eqref{eq:solve:this:K} can not be transformed into its more efficient normal form and solved with the SVD, but with standard linear least-square techniques instead.
However, more important is the fact that the equation is independent of $\R{X}$ and $\R{B}$.
It is therefore sufficient to measure or estimate the position $\t{B}$, thereby reducing the calibration object to a single point.
As with most Kronecker product approaches, the retrieved rotation $\R{Y}$ needs to be reorthonormalized.
Therefore, $\t{Y}$ is recomputed again to compensate for the error correction.

\begin{figure*}[!ht]
\centering
\subfigure[][Rotation error]{
	\includegraphics[width=0.3\textwidth]{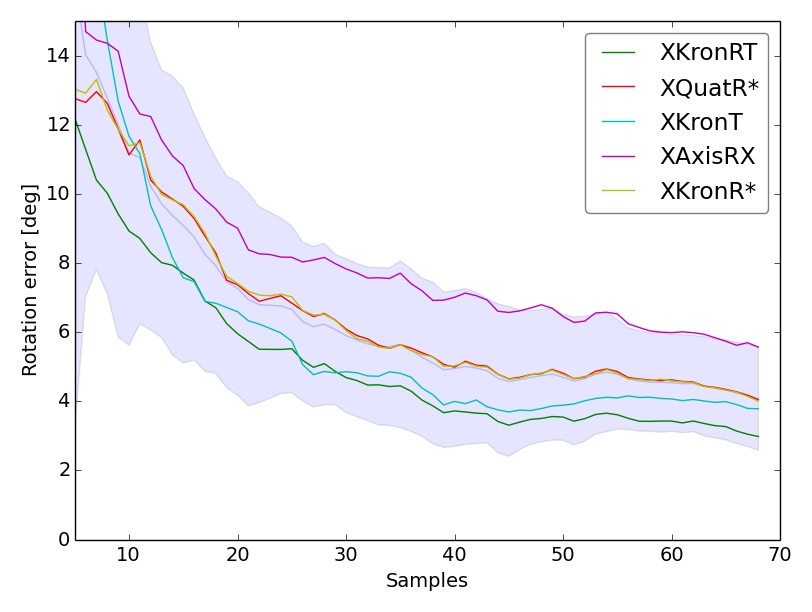}
	\label{fig:AXXB_R}
}
\subfigure[][Translation error]{
	\includegraphics[width=0.3\textwidth]{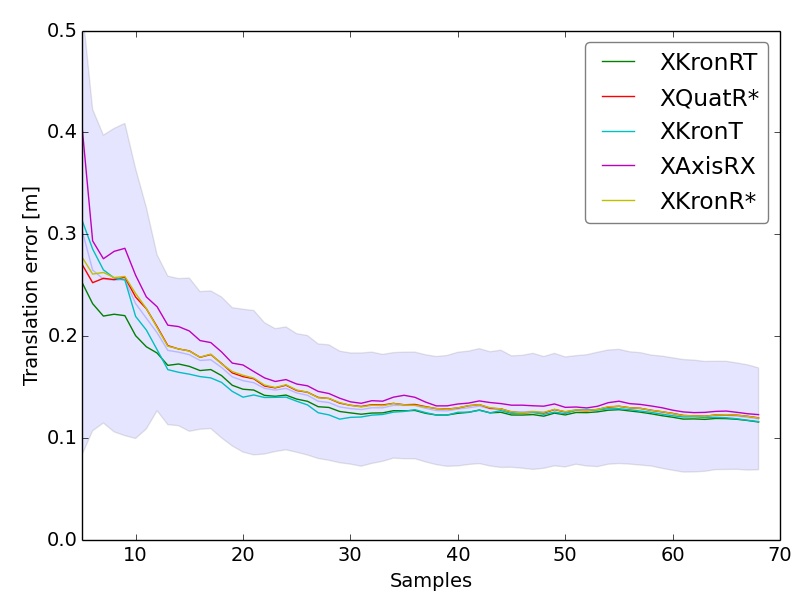}
	\label{fig:AXXB_T}
}
\subfigure[][Reprojection error]{
	\includegraphics[width=0.3\textwidth]{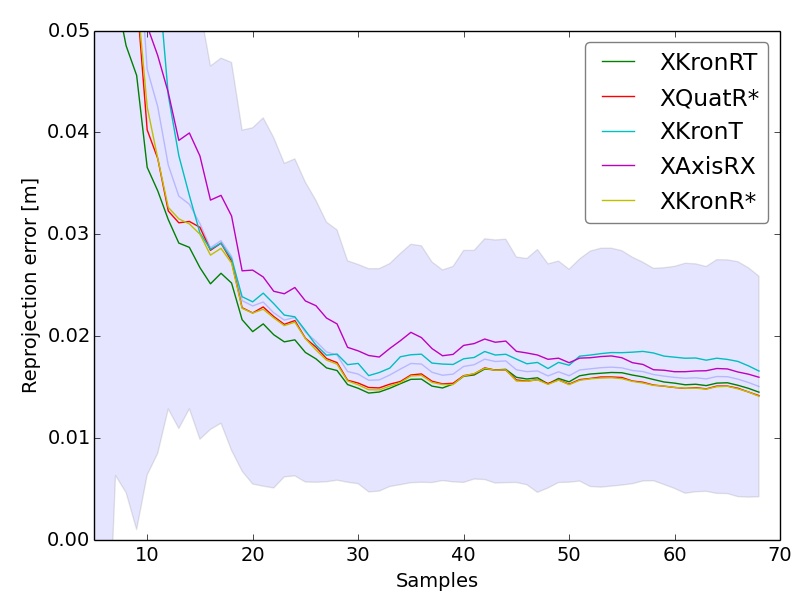}
	\label{fig:AXXB_P}
}
\subfigure[][Rotation error]{
	\includegraphics[width=0.3\textwidth]{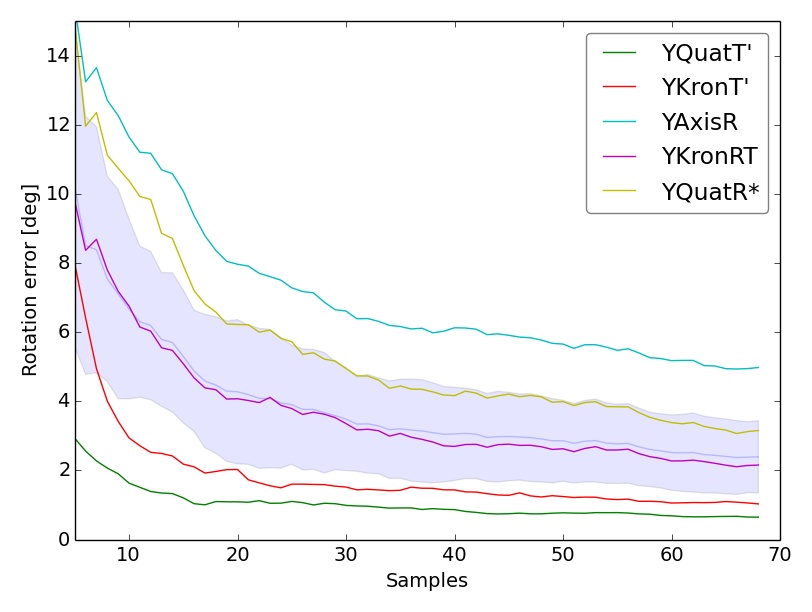}
	\label{fig:AXYB_R}
}
\subfigure[][Translation error]{
	\includegraphics[width=0.3\textwidth]{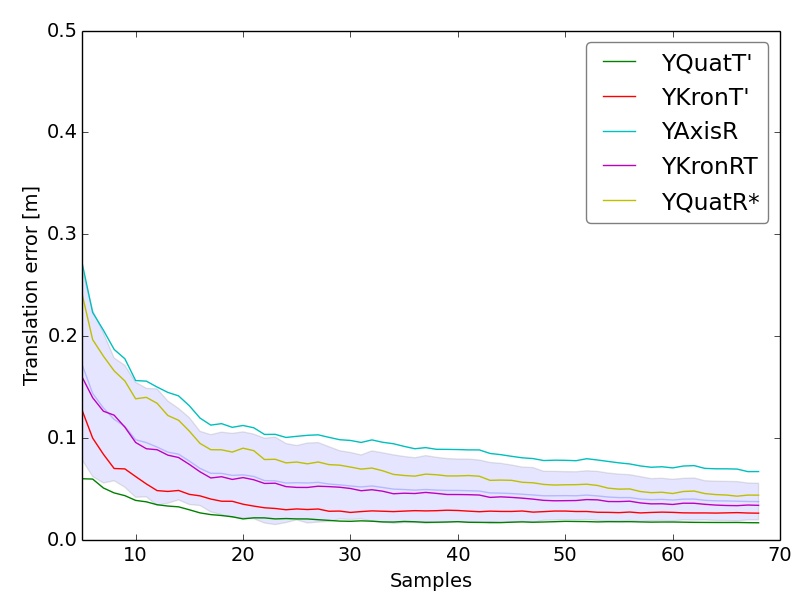}
	\label{fig:AXYB_T}
}
\subfigure[][Reprojection error]{
	\includegraphics[width=0.3\textwidth]{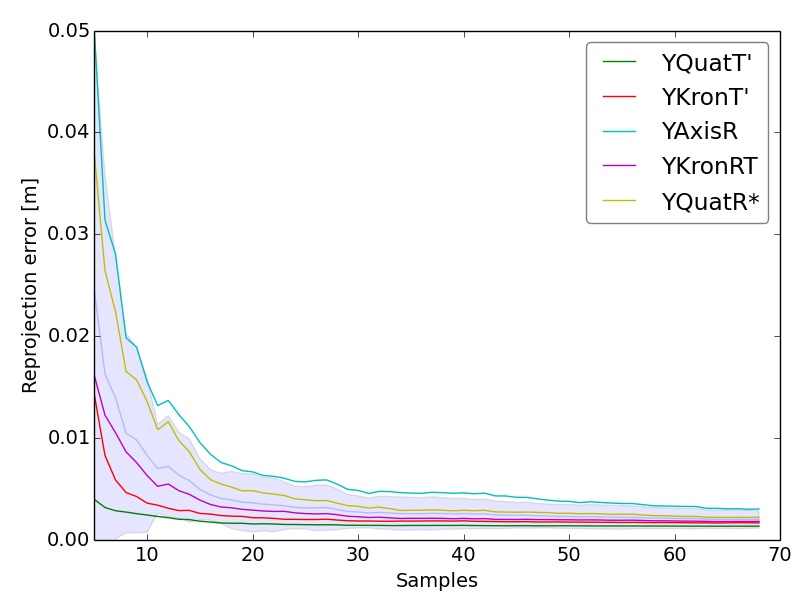}
	\label{fig:AXYB_P}
}
\caption{Comparison of absolute errors for (top) AX=XB and (bottom) AX=YB  with average mean and standard deviation in grey with increasing number of samples.}
\vspace{40pt}
\label{fig:compare}
\end{figure*}

\begin{figure*}[!ht]
\centering
\subfigure[][Rotation error]{
	\includegraphics[width=0.3\textwidth]{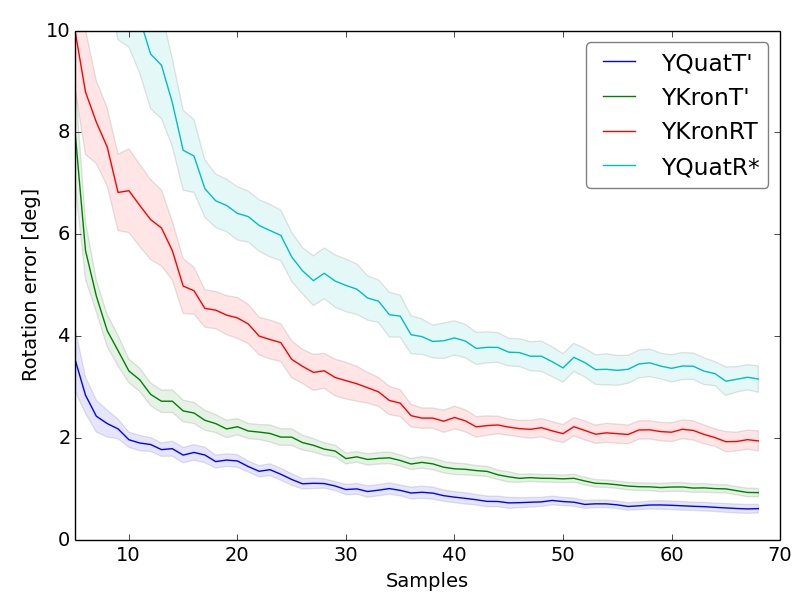}
	\label{fig:YBest_R}
}
\subfigure[][Translation error]{
	\includegraphics[width=0.3\textwidth]{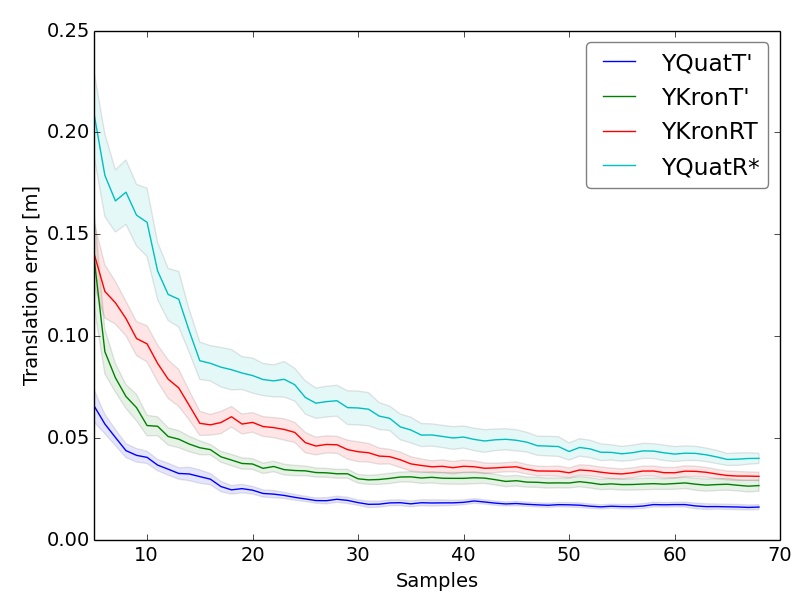}
	\label{fig:YBest_T}
}
\subfigure[][Reprojection error]{
	\includegraphics[width=0.3\textwidth]{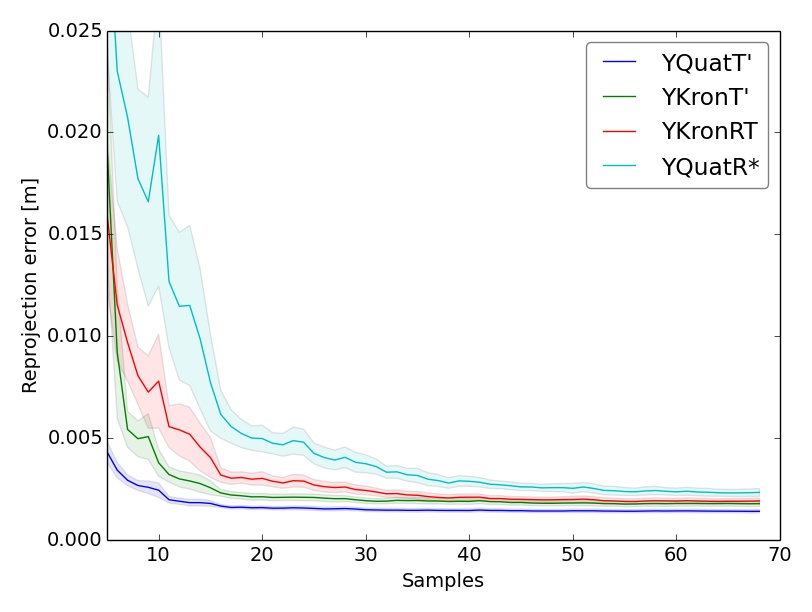}
	\label{fig:YBest_P}
}
\caption{Comparison of the absolute errors with 20\% standard deviation of the 4 best calibration methods with increasing number of samples. }
%\vspace{-10pt}
\label{fig:YBest}
\end{figure*}

\begin{figure*}[!ht]
\centering
\subfigure[][Rotation error]{
	\includegraphics[width=0.3\textwidth]{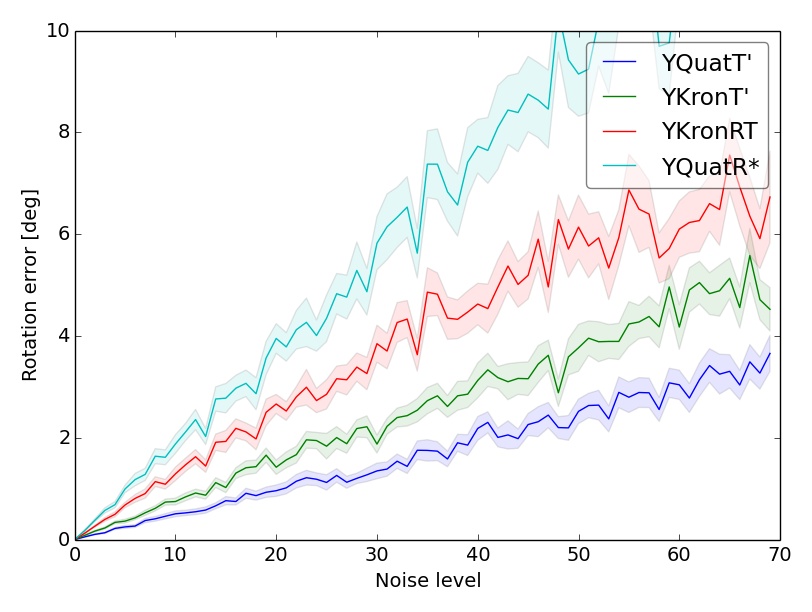}
	\label{fig:YBestNoise_R}
}
\subfigure[][Translation error]{
	\includegraphics[width=0.3\textwidth]{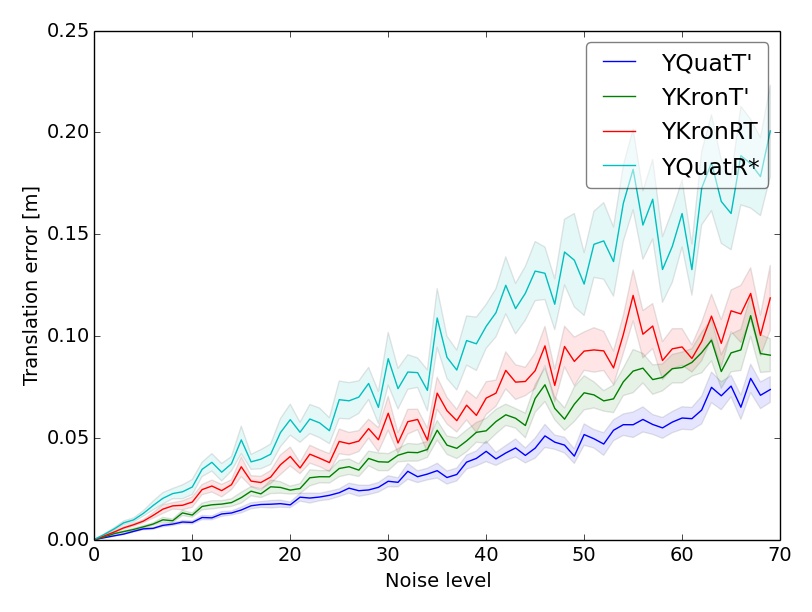}
	\label{fig:YBestNoise_T}
}
\subfigure[][Reprojection error]{
	\includegraphics[width=0.3\textwidth]{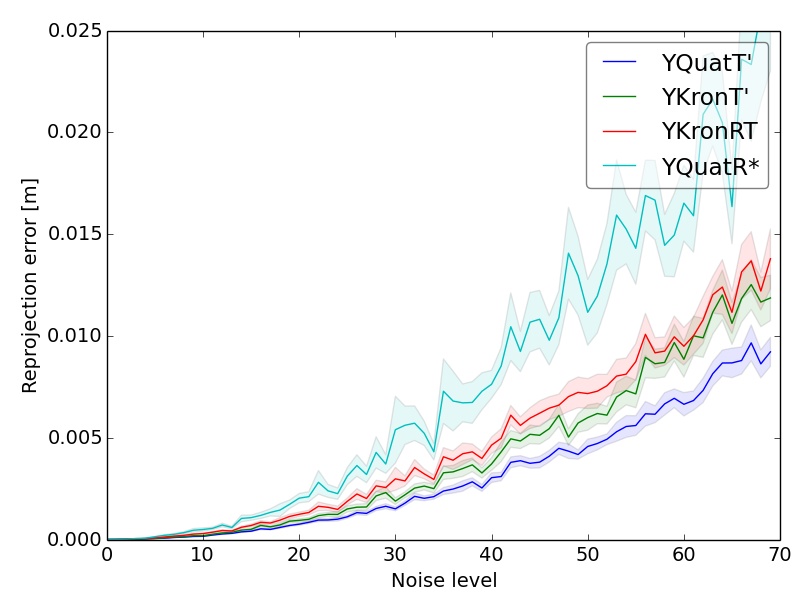}
	\label{fig:YBestNoise_P}
}
\caption{Comparison of the absolute errors with 20\% standard deviation of the 4 best calibration methods with increasing noise level. }
\vspace{10pt}
\label{fig:YBestNoise}
\end{figure*}

\section{EXPERIMENTS}\label{sec:experiments}

The evaluation of the new approaches has been done in comparison to the implementations derived from the general solution \ref{sec:solutions}.
Preliminary test showed that not all of these algorithms behave stably, in particular formulas based on the equation of translation in quaternion representation due to its double cover of the rotation group $SO(3)$ (cf. \cite{1998-Dornaika}).
Additionally, some of the solutions show a similar behavior such that they can be represented by only one implementation.
Therefore, we restrict the evaluation to a selection as shown in Tbl.~\ref{tbl:implementations}.
The evaluation has been conducted with simulated data.

%\begin{table}[htb]
%%	\renewcommand{\arraystretch}{1.5}
%	\centering
%	\resizebox{\columnwidth}{!}{
%	{\tiny
%	\begin{tabular}{@{}lllll@{}}
%	 	\toprule
%	 	\thead{TYPE} & \thead{REPR.} & \thead{EQ.} & \thead{AUTHOR} & \thead{NAME} \\
%	 	\midrule
%	 	AX=XB & AxisAngle & R  (\ref{eq:solve:R:2}) & Park~\cite{1994-Park} & XAxisR \\
%	 	AX=XB & Kronecker & R  (\ref{eq:solve:R:3}) & cf. Shah~\cite{2013-Shah} & XKronR \\
%	 	AX=XB & Quaternion& R  (\ref{eq:solve:R:3}) & cf. Dornaika~\cite{1998-Dornaika} & XQuatR \\
%	 	AX=XB & Kronecker & RT (\ref{eq:solve:R:1},\ref{eq:solve:t}) & Andreff~\cite{2001-Andreff} & XKronRT \\
%	 	AX=XB & Kronecker & T  (\ref{eq:solve:t}) & cf. \textit{this} & XKronT' \\
%	 	\midrule
%	 	AX=YB & AxisAngle & R  (\ref{eq:solve:R:1}) & Zhuang~\cite{1994-Zhuang} & YAxisR \\
%	 	AX=YB & Quaternion& R  (\ref{eq:solve:R:3}) & Dornaika~\cite{1998-Dornaika} & YQuatR \\
%	 	AX=YB & Kronecker & RT (\ref{eq:solve:R:1}), \ref{eq:solve:t} & Li~\cite{2010-Li} & YKronRT \\
%	 	AX=YB & Kronecker & T  (\ref{eq:solve:t}) & \textit{this} & YKronR' \\
%	 	AX=YB & Quaternion& T  (\ref{eq:solve:t}) & \textit{this} & YQuatT' \\
%		\bottomrule
%	\end{tabular}
%	}}
%	\caption{Blabla}
%	\label{tab:1}
%\end{table}

\begin{figure*}[!ht]
\centering
\subfigure[][Rotation error]{
	\includegraphics[width=0.4\textwidth]{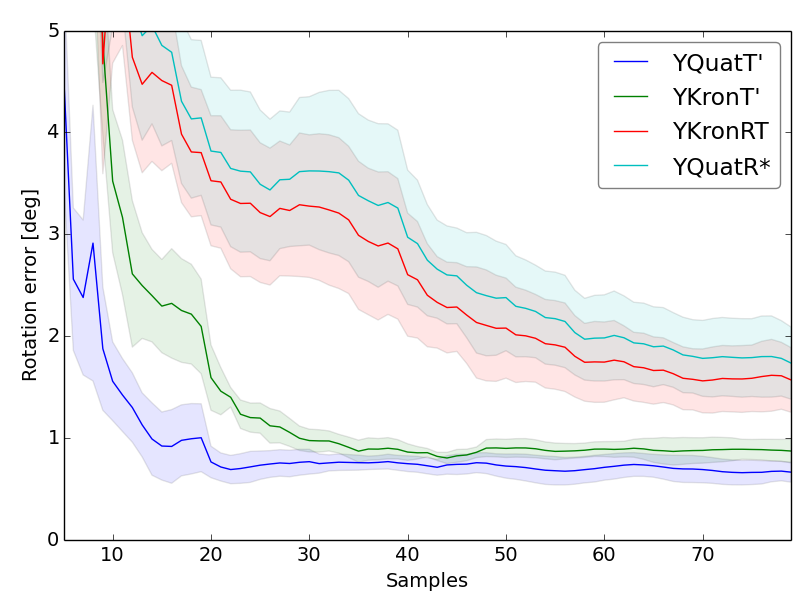}
	\label{fig:Real1_R}
}
\subfigure[][Translation error]{
	\includegraphics[width=0.4\textwidth]{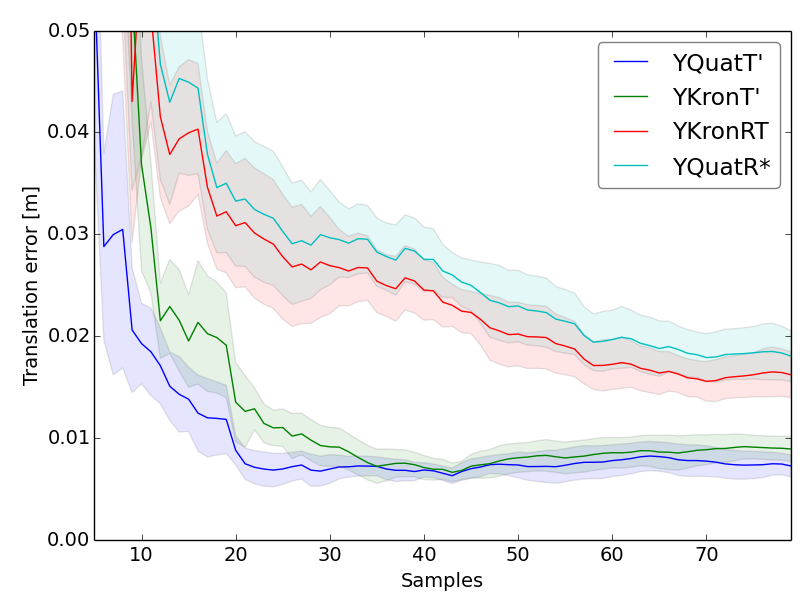}
	\label{fig:Real1_T}
}
\caption{Comparison of the absolute errors with 50\% standard deviation of the 4 best calibration methods in a real eye-on-hand calibration setup. }
%\vspace{-10pt}
\label{fig:Real1}
\end{figure*}

\subsection{Simulation}
In the simulation, we use the following setup:
For each round, $\X = \T[E]{M}$, $\A = \T[R]{E}$ and $\B = \T[C]{E}$ are chosen randomly, but constrained to the distance intervals $\M[\empty]{t}{X} = 0.2\text{m} \pm 0.1\text{m}$, $\M[\empty]{t}{A} = 1.25\text{m} \pm 0.75\text{m}$ and $\M[\empty]{t}{B} = 0.8\text{m} \pm 0.3\text{m}$ with $\Y = \A \X \B^{-1}$ to simulate a realistic setup.
We keep the ground truth $\X$ and $\Y$ and generate random samples for $\A$ in the same distance interval, but now with a restricted orientation within the range $\pm 30^{\circ}$; $\B$ thereby becomes $\B = \Y^{-1} \A \X$. 
$\A$ and $\B$ are then multiplied by transformations $\M[A]{N}{\empty}$ and $\M[B]{N}{\empty}$, respectively, to simulate measurement and sensor noise which is chosen according to the experiment.
Each experiment is done in multiple rounds from which we then compute the mean rotation, translation and squared reprojection error.

\subsubsection{Experiment 1 -- AX=XB vs AX=YB}\ \newline
In our first two experiments, we compare accuracy, robustness and convergence behavior of algorithms based on the used equations given a static noise $|\M[A]{N}{\alpha}| < 15^{\circ}$, $|\M[A]{N}{t}| < 0.02\text{m}$ and $|\M[B]{N}{\alpha}| < 10^{\circ}$, $|\M[B]{N}{t}| < 0.01\text{m}$.
Here, we compare the algorithms based on the formula $AX=XB$ and $AX=YB$ with increasing sample size up to 70 samples.
For each added sample, each algorithm is run for 30 rounds and its mean error is computed.
We furthermore compute the average standard deviation of each equation type which directly correlate to the robustness of the approaches.

\figref{fig:compare} shows the results of the experiments with the $AX=XB$ methods in the top row and the $AX=YB$ methods in the bottom row.
The average standard deviation $\pm 1\sigma$ from the average mean for each equation type is marked as a grey area.

The experiment shows that most of the $AX=XB$ algorithms behave very similarly and have a moderate standard deviation for the rotational and translational error.
From the reprojection error, however, we can conclude that the quality of the estimation in terms of robustness varies up $50\%$ which is an undesirable behavior.
The $AX=YB$ algorithms, in contrast, show a very robust behavior, converging faster with a more accurate estimation.
Only the rotation error varies up to a couple of degrees depending on the approach which we analyze in more detail in the next experiment.

\subsubsection{Experiment 2 -- Best of AX=YB}\ \newline
We are using the same setup as in experiment 1, but this time only for the 4 best performing algorithms of type $AX=YB$ which can be clearly identified from \figref{fig:compare} as \textbf{YQuatT'}, \textbf{YKronT'}, \textbf{YQuatR*} and finally \textbf{YKronRT}.

\figref{fig:YBest} shows a closer view on the different methods, this time independently with their respective standard deviation of $\pm 0.2\sigma$.

We furthermore run an additional test to evaluate the behavior under increasing noise conditions.
Therefore, we increase the rotation noise linearly in 70 steps from $0^{\circ}$ to $30^{\circ}$ and $20^{\circ}$ respectively for $\M[A]{N}{\alpha}$ and $\M[B]{N}{\alpha}$, while the translation error increases at the same time from $0\text{m}$ to $0.06\text{m}$ and $0.03\text{m}$ respectively for $\M[A]{N}{t}$ and $\M[B]{N}{t}$.
For each iteration, we compute the mean and variance error of 40 rounds.

\figref{fig:YBestNoise} depicts the behavior of each algorithm with increasing noise along with their respective standard deviation of $\pm 0.2\sigma$.

The experiments clearly show the difference in terms of accuracy, convergence and robustness between the different methods.
Both our proposed algorithms, \textbf{YQuatT'} and \textbf{YKronT'}, converge faster than the state-of-the-art algorithms \textbf{YQuatR*} and \textbf{YKronRT} and show a slightly better performance in terms of robustness.
In terms of accuracy, our solutions surpass the others as well, especially the rotation error is magnitudes lower.
Interestingly, our quaternion-based approach \textbf{YQuatT'} gives the best results whereas in most other implementations, the Kronecker product outperforms the quaternion version.
Furthermore, we can see that our proposed algorithm \textbf{YKronT'} which only uses point measurements from the calibration object, only performs slightly worse than the best method \textbf{YQuatT'} and still better than the best state-of-the-art algorithm \textbf{YKronRT}.

\subsection{Real world}
For the evaluation on a real robotic system, we use a Xtion RGB-D camera mounted on a rack above a UR10 robotic arm.
Our calibration object is a $5cm\times5cm$ Aruco marker taped on the endeffector.
The marker pose is used by all algorithms except for the proposed \textbf{YKronT'} which solely uses the center position of the marker.
For the experiment, we place the robotic arm about $0.7\text{m}$ away from the camera and then move the arm continuously in random poses within $\pm 0.15\text{m}$ in each direction and an angular deviation of $\pm 40^{\circ}$.
The mean and standard deviation is computed over 5 rounds with each 80 samples.
Since we only estimate $\Y$ and $\X$ remains unknown, we cannot compute the reprojection error.
The angular and translational error are compared against a manually refined calibration, therefore we can mainly observe the behavior in terms of robustness.

\figref{fig:Real1} shows the result of the 4 best algorithms for the real setup.
Due to the small marker size used in the experiment, the orientation estimation of the marker is usually much higher compared to the position estimation.
As expected from the simulated experiments, we can thus see that the methods based on the equation of translation perform in general much better than the sequential and simultaneous methods.
Both proposed methods \textbf{YQuatT'} and \textbf{YKronT'} show a much faster convergence rate and are also more robust in their estimation.
The quaternion approach gives still the best results, closely followed by the point-based Kronecker approach.

In conclusion, the experiments showed that the proposed solution works robustly and estimate the transformation with high accuracy.
Especially the \textbf{YKronT'}, which uses less information compared to the others, shows a very desirable behavior and can therefore be used for a continuous calibration process.

%
%\subsection{Real world}
%
%\begin{itemize}
%\item Marker size?
%\item Robustness (repeatability)
%\item Quaternion vs Kroncker
%\item Simultaneous vs 2 step?
%\item Convergence (measurements)
%\end{itemize}

\section{CONCLUSION}\label{sec:conclusion}

In this paper, we aimed to tackle two seemingly contradicting developments in the research of solving the calibration problem:
On the one hand, mathematical formulations have been proposed to increase the accuracy of the calibration through linear and non-linear methods.
However, the accuracy of the pose estimation of the calibration object has a huge impact on the quality of the solutions and therefore gives better results for larger objects.
On the other hand, research towards self-calibration has been conducted which makes the use of calibration objects obsolete and is replaced by feature matching or other estimation methods which in turn decrease the accuracy of the solutions.

To solve this dilemma, we first analyze known closed-form solutions and rephrase them in an more general way such that we can derive a structured and systematic view on the possible solution space.
We can thereby show that known methods are part of this solution space, but it also gives rise to new possible approaches.
In particular, we conclude that the focus for solving the hand-eye problem mainly lies on the equation of rotation, sometimes simultaneously used in combination with the equation of translation.

Therefore, this paper proposes a shift towards the equation of translation only and shows that the calibration problem can, in fact, still be solved.
We combine this method with the assumption that the pose of the calibration object is irrelevant to the calibration problem and hence has not to be solved for.
This leads to a new formulation which surpasses other closed-form solutions in terms of accuracy, robustness and convergence as concluded from the conducted experiments.

Finally, we formulate a method based on the equation of translation with similar performance and quality, but does not need the estimated orientation of the calibration object to solve the calibration problem and hence can be reduced to a single 3D point.
This renders our proposed method an accurate and flexible solution which allows continuous hand-eye calibration.

%\addtolength{\textheight}{-10cm}   % This command serves to balance the column lengths
\balance

\section*{ACKNOWLEDGMENT}
The research leading to these results has received funding from the
European Commission's 7th Framework Programme under grant agreement
no.~610917 (STAMINA).

%%% References
\bibliographystyle{IEEEtran}
\bibliography{bibliography}

\end{document}